\documentclass[runningheads]{llncs}

\usepackage{times}
\usepackage{epsfig}
\usepackage{graphicx}
\usepackage{amsmath}
\usepackage{amssymb}
\usepackage{booktabs}
\usepackage{color, colortbl}
\usepackage[dvipsnames]{xcolor}
\usepackage{subcaption}
\usepackage{arydshln}
\usepackage{mathtools}
\usepackage{amsmath}

\usepackage{tikz}
\usepackage{comment}
\usepackage[accsupp]{axessibility}  % Improves PDF readability for those with disabilities.

\DeclareCaptionLabelFormat{andtable}{#1~#2  \&  \tablename~\thetable}
\definecolor{Res7}{rgb}{1.0,1.0,1.0}
\definecolor{Res6}{rgb}{0.93,0.93,0.93}
\definecolor{Res5}{rgb}{0.99,0.94,0.91}
\definecolor{Res3}{rgb}{0.93,0.96,0.91}
\definecolor{Res4}{rgb}{0.93,0.95,0.98}
\definecolor{Res2}{rgb}{0.95,0.95,0.95}
\definecolor{Res1}{rgb}{1.0,0.97,0.90}
\DeclareRobustCommand\onedot{\futurelet\@let@token\@onedot}
\def\@onedot{\ifx\@let@token.\else.\null\fi\xspace}

\newcommand{\figref}[1]{Fig.~\ref{#1}}
\newcommand{\tabref}[1]{Tab.~\ref{#1}}
\newcommand{\secref}[1]{Sec.~\ref{#1}}

\newcommand{\eqrefhc}[1]{Eq.~(\ref{#1})}

\usepackage{color}
\definecolor{light}{rgb}{0.6, 0.6, 0.6}

\newcommand{\dashrule}[1][black]{%
  \color{#1}\rule[\dimexpr.5ex-.2pt]{4pt}{.4pt}\xleaders\hbox{\rule{4pt}{0pt}\rule[\dimexpr.5ex-.2pt]{4pt}{.4pt}}\hfill\kern0pt%
}
\newcommand{\rulecolor}[1]{%
  \def\CT@arc@{\color{#1}}%
}

\newenvironment{conditions*}
  {\par\vspace{\abovedisplayskip}\noindent
   \tabularx{\columnwidth}{>{$}l<{$} @{}>{${}}c<{{}$}@{} >{\raggedright\arraybackslash}X}}
  {\endtabularx\par\vspace{\belowdisplayskip}}

\usepackage{hyperref}  %hyperref still needs to be put at the end!

\begin{document}
% \renewcommand\thelinenumber{\color[rgb]{0.2,0.5,0.8}\normalfont\sffamily\scriptsize\arabic{linenumber}\color[rgb]{0,0,0}}
% \renewcommand\makeLineNumber {\hss\thelinenumber\ \hspace{6mm} \rlap{\hskip\textwidth\ \hspace{6.5mm}\thelinenumber}}
% \linenumbers
\pagestyle{headings}
\mainmatter
\def\ECCVSubNumber{4208}  % Insert your submission number here

\title{GOCA: Guided Online Cluster Assignment \\ for Self-Supervised Video Representation Learning} 

% INITIAL SUBMISSION 
\begin{comment}
% \titlerunning{ECCV-22 \ECCVSubNumber} 
% \authorrunning{ECCV-22 \ECCVSubNumber} 
% \author{Anonymous ECCV submission}
% \institute{Paper ID \ECCVSubNumber}
\end{comment}
%******************

% CAMERA READY SUBMISSION
%\begin{comment}
\titlerunning{GOCA}
% If the paper title is too long for the running head, you can set
% an abbreviated paper title here
%
\author{Huseyin Coskun\thanks{Work done during internship at Snap Inc.}\inst{1, 2} \and Alireza Zareian\inst{1}
\and Joshua L. Moore\inst{1}
\and \newline Federico Tombari\inst{2,3}
\and  Chen Wang\inst{1}
}
\authorrunning{ }
% First names are abbreviated in the running head.
% If there are more than two authors, 'et al.' is used.
%
\institute{Snap Inc.  
\and
TU Munich, Germany
\and
Google
\\
\email{hcoskun@snap.com}}

%\end{comment}
%******************
\maketitle
\begin{abstract}

Clustering is a ubiquitous tool in unsupervised learning. Most of the existing self-supervised representation learning methods typically cluster samples based on visually dominant features. While this works well for image-based self-supervision, it often fails for videos, which require understanding motion rather than focusing on background. Using optical flow as complementary information to RGB can alleviate this problem. However, we observe that a na\"ive combination of the two views does not provide meaningful gains. In this paper, we propose a principled way to combine two views.
Specifically, we propose a novel clustering strategy where we use the initial cluster assignment of each view as prior to guide the final cluster assignment of the other view.
% Specifically, we propose a novel clustering strategy where we use each modality to guide cluster assignment by using initial assignments as priors to each other.
This idea will enforce similar cluster structures for both views, and the formed clusters will be semantically abstract and robust to noisy inputs coming from each individual view. Additionally, we propose a novel regularization strategy to address the feature collapse problem, which is common in cluster-based self-supervised learning methods. Our extensive evaluation shows the effectiveness of our learned representations on downstream tasks, e.g., video retrieval and action recognition. Specifically, we outperform the state of the art by 7\% on UCF and 4\% on HMDB  for video retrieval, and 5\% on UCF and 6\% on HMDB for video classification. 
\footnote[1]{Code available at https://github.com/Seleucia/goca}
  
%   \cHu{Grammar needs to be checked. Also would it make sense to mention Prototype Regularization in abstract?  }

%   . Extensive experiments demonstrate that
% FAME can significantly boost the performance in different
% downstream tasks with various backbones. When integrated
% with MoCo, FAME reaches 84.8% and 53.5% accuracy on
% UCF101 and HMDB51, respectively, achieving the state-ofthe-art performance.

%   However, these pseudo labels are noisy
% even with consistency check or confidence-based filtering
% due to the domain shift in the data. To solve this problem, we design a multi-scale domain adaptation module
% (MDAM) to reduce the domain gap between the synthetic
% and real data.
\end{abstract}

%%%%%%%%% BODY TEXT
\section{Introduction} 

\begin{figure}[ht]
\centering
 \includegraphics[trim=50 120 90 90,width=0.65\linewidth]{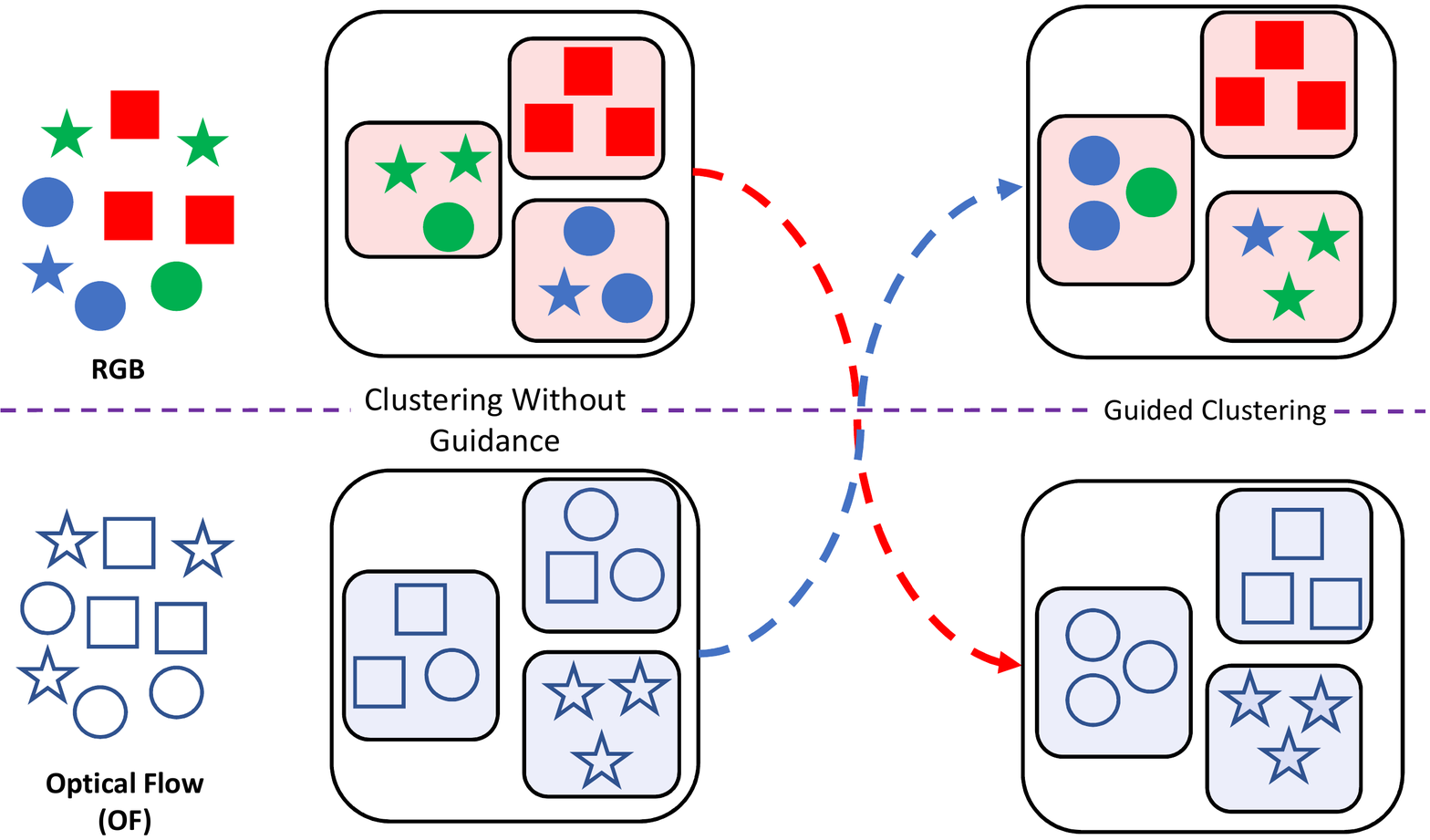}
\caption{An abstract illustration of the proposed idea. The filled and empty shapes represent \emph{RGB} and \emph{OF} views respectively. Each shape represents a different type of activity, while colors represent intra-class variations. In this toy example, the initial clustering is wrong for both views. \emph{RGB} fails because it clusters based on color (e.g. irrelevant background information), while \emph{OF} fails since it confuses square with circle (due to the lower resolution of details). By combining the two types of information, clustering can be achieved correctly on both views.}
\label{fig:teaser}
\end{figure}

The pursuit of understanding human activities in videos is a fundamental problem in computer vision. Representation learning methods with supervised training strategies showed promising results on various tasks, such as action understanding~\cite{carreira2017quo,Coskun21,feichtenhofer2019slowfast,jiang2015human,liu2016video}, action detection and localization~\cite{chao2018rethinking,zhao2017temporal}, and action proposal~\cite{gao2018ctap,liu2019multi}. It is fair to say that large-scale, manually labeled video datasets such as Kinetics~\cite{kay2017kinetics}, AVA~\cite{gu2018ava}, and Epic~\cite{damen2018scaling} substantially contributed to that success. In spite of those promising results, these algorithms can usually only recognise activities if they have access to a semantically labelled dataset. The cost and challenges of collecting large-scale, manually labelled videos hinder further improvements in activity understanding. On the other hand, the internet is a virtually unlimited source of unlabeled videos (e.g. YouTube). Therefore, designing a representation learning strategy that does not rely on manual labelling is fundamentally important.

Self-supervised learning (SSL) aims to address this issue, by designing pretext 
tasks that only rely on input, and training networks to solve those tasks. Recent advances in SSL for image understanding~\cite{caron2020unsupervised,chen2020simple,NEURIPS2020_f3ada80d,han2020memory,he2020momentum,richemond2020byol,tian2020contrastive,zhuang2019local} have achieved excellent performance in various downstream tasks. Motivated by this success, several papers brought these ideas to the video domain~\cite{dave2021tclr,feichtenhofer2021large,qian2021spatiotemporal,sun2019learning}. Although these methods show promising results to some extent, they rely solely on an \emph{RGB} stream. As demonstrated by ~\cite{alwassel2020self,asano2020labelling,Han20}, this is not sufficient to learn a strong temporal representation.

Johansson's classical psychology work~\cite{johansson1973visual} shows that humans can recognize activities by only watching a few bright dots depicting the movement of the main body joints. This intuitive work motivated researchers to use optical flow as a representation of motion for activity understanding~\cite{carreira2017quo,feichtenhofer2016convolutional,sevilla2018integration,simonyan2014two}, and they have achieved significant improvements over \emph{RGB}-only models in the supervised learning literature. 
Inspired by this success, many recent SSL works~\cite{gavrilyuk2021motion,Han20,toering2021selfsupervised}
have explored using optical flow (\emph{OF}) to advance SSL beyond \emph{RGB}-only baselines. Han~\emph{et al.} (CoCLR)~\cite{Han20} used \emph{OF} to retrieve positive samples for the infoNCA~\cite{oord2018representation}, which led to significant improvements. Nevertheless, CoCLR did not utilize \emph{OF} for training the backbone and hence may not have realized the full potential of learning motion representations. VICC~\cite{toering2021selfsupervised} adopted the online cluster assignment~\cite{caron2020unsupervised} to videos by considering \emph{OF} as another view of \emph{RGB}, and minimized the distance between \emph{RGB} and \emph{OF} features during online clustering. This method obtained SOTA results on various datasets. However, enforcing similarity between \emph{RGB} and \emph{OF} features can be detrimental, especially if one information source is noisy, as is usually the case for OF due to camera motion. Furthermore, these models~\cite{Han20,toering2021selfsupervised} require a complicated training strategy that successively updates one model while freezing the parameters of the other model, which prevents end-to-end training.
% , furthermore since two rgb and flow models trained separately, there if no 

In this paper, we introduce the guided online cluster assignment algorithm (GOCA) to address the aforementioned problems. Specifically, for a given video with \emph{RGB} and \emph{OF} representations, we first compute initial cluster assignments for only using \emph{RGB} or \emph{OF} separately, and then we use these assignments as priors for each other to compute a final assignment that is guided by both views, as illustrated in \figref{fig:teaser}. After we obtain our final assignment, we train a backbone network by minimizing a cross entropy loss between the final cluster assignment of different augmentations of the same video (as used in ~\cite{caron2020unsupervised}). The proposed idea has several benefits compared to the state of the art~\cite{caron2020unsupervised,Han20,toering2021selfsupervised}. First, it constructs more robust clusters during training due to prior information, which is particularly important when one information source is noisy. Second, allowing \emph{RGB} and \emph{OF} to share information by means of sharing cluster assignment encourages the two views to form similar cluster structure, which leads to more semantically abstract representations. Third, both \emph{RGB} and \emph{OF} backbones are trained jointly and information flows both ways during training, which is beneficial for both backbones due to the complementary nature of these views. Fourth, compared to the CoCLR method~\cite{Han20}, \emph{OF} is utilized more explicitly in our formulation, which leads to stronger spatio-temporal representations. Finally, the proposed approach circumvents complicated training strategies~\cite{Han20,gavrilyuk2021motion,toering2021selfsupervised} and allows simple and end to end training. 

We also propose a novel prototype regularization method to address the feature collapse problem, where all features are mapped to a single point. This is a common problem in SSL and was partially addressed in SwAV~\cite{caron2020unsupervised} via equipartition
constraint. However, it requires careful tuning\footnote[1]{https://github.com/facebookresearch/swav\#common-issues} of the parameter $\lambda$ (See \eqrefhc{eq:reg_ot}) of the Sinkhorn algorithm, which makes it hard used in practice~\cite{cai2021theory,chen2021exploring,chen2021empirical,regatti2021consensus}. We address this problem by constructing cluster prototypes which are maximally distant from each other. We achieve this by locating the $N$ prototypes in the $\Phi$ dimensional space such that they divide the space equally. Despite the simplicity of the proposed idea, it yields consistent performance improvements.

To sum up, our contributions are:
1) we introduce a novel Guided Online Cluster Assignment (GOCA) algorithm that aims to learn stronger spatio-temporal representations by utilizing the complementary information of \emph{RGB} and \emph{OF}; 2) we mathematically prove that guidance based clustering can be achieved efficiently with the Sinkhorn algorithm; 3) we propose a prototype regularization strategy that addresses the common feature collapse problem; 4) we perform extensive evaluations of our method using two backbones (S3D~\cite{xie2018rethinking} and R(2+1)D~\cite{tran2018closer}) on four different evaluation regimes (See \secref{sec:experiments}). 
The proposed model outperforms the state of the art in almost all experiments. Furthermore, we present ablation studies to show the effect of each contribution and the key parameters of our method.
% \newpage

\begin{figure*}[ht]
\centering
 \includegraphics[trim=100 270 130 120,width=0.65\linewidth]{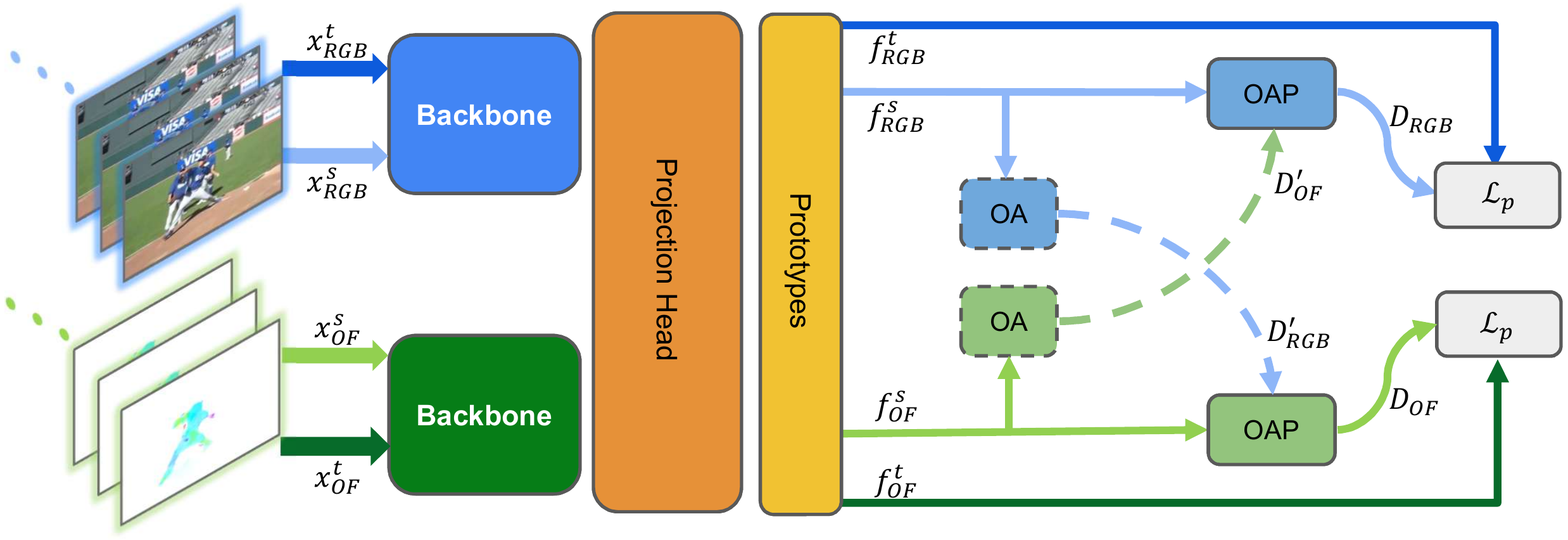}
\caption{An overview of the proposed GOCA algorithm. The \emph{RGB} and Optical Flow (\emph{OF}) views have the same backbone architectures and share the projection head and prototypes. 
Optimal assignment (OA) and optimal assignment with prior (OAP) are defined in  \eqrefhc{eq:prior_ot_rgb} and \eqrefhc{eq:prior_ot_flow}, respectively. $\mathcal{L}_{p}$ is defined for a batch of videos in \eqrefhc{eq:curly_l_p}} \label{fig:architecture}
\end{figure*}
\section{Related Works}\label{sec:related}
\noindent
\paragraph{Self-supervised representation learning.}

The success of image-based SSL methods ~\cite{doersch2015unsupervised,dosovitskiy2015discriminative,he2020momentum,larsson2016learning,noroozi2016unsupervised,zhuang2019local,richemond2020byol,tian2020contrastive,zhang2016colorful} has inspired their application in the video domain, resulting in several recently proposed methods in that domain.
% The success of image based self-supervised learning methods~\cite{noroozi2016unsupervised,doersch2015unsupervised,zhang2016colorful,larsson2016learning,dosovitskiy2015discriminative,zhuang2019local,he2020momentum,tian2020contrastive,richemond2020byol} brings attention to videos, hence recently several methods were proposed for videos. 
Early approaches for video representation learning~\cite{ahsan2018discrimnet,diba2019dynamonet,srivastava2015unsupervised,vondrick2016generating} typically relied on the idea of predicting the future frames given the past frames, which requires carefully training a deep generative model. Other works focus on designing proxy tasks to exploit temporal information. In a pioneering work, Misra~\emph{et al.}~\cite{misra2016shuffle} designed a pretext task that predicts whether the order of video frames is correct or wrong. 
Follow-up methods achieved better performance by designing new pretext tasks such as predicting clip order~\cite{fernando2017self,lee2017unsupervised,misra2016shuffle,xu2019self}, pace ~\cite{benaim2020speednet,cho2020self,wang2020self,yao2020video}, or the arrow of time ~\cite{pickup2014seeing,wei2018learning}. To achieve better generalization, aside from designing novel pretext tasks, some recent works focus on using instance-based contrastive learning approaches~\cite{dave2021tclr,feichtenhofer2021large,Hu_2021_ICCV,lin2021self,pan2021videomoco,qian2021spatiotemporal}. Since videos often contain multi-modal data, many recent works used audio \cite{akbari2021vatt,alayrac2020self,alwassel2020self,alwassel_2020_xdc,asano2020labelling,NEURIPS2018_c4616f5a,patrick2020multi} or language \cite{akbari2021vatt,alayrac2020self,miech2020end,sun2019videobert} as complementary information to improve results. Unlike those works, we focus on learning representations purely on visual input, namely \emph{RGB} and optical flow (\emph{OF}). Note that we compute \emph{OF} from \emph{RGB} videos in an unsupervised way. 

\noindent
\paragraph{Clustering based self-supervised learning.}
Early works in this direction adopted clustering to representation learning by simply applying clustering to obtain pseudo-labels~\cite{alwassel2020self,caron2018deep,yang2016joint}. The common approach has two alternating steps, as it performs clustering offline, rather than concurrently with training as described in \secref{sec:preliminaries}. 
One of the main limitations of that approach is that such naive implementations may form degenerate clusters where all samples are clustered to the same point. Asano~\emph{et al.}~\cite{asano2020labelling,asano2020self} tackle this problem by adding an equipartition constraint on the number of samples per cluster and converting the pseudo-label generation step into an optimal assignment problem. They solve the resulting problem using the Sinkhorn-Knopp algorithm~\cite{cuturi2013sinkhorn}. Although the equipartition constraint achieves promising results, it is still inefficient due to the offline clustering. Recently, Caron~\emph{et al.}~\cite{caron2020unsupervised} (closest work to ours) eliminated offline label generation via an online clustering algorithm which allows them to train end-to-end on large-scale datasets. Although it works well for images, we observe that adapting this idea to videos is not as effective as with images. As seen in ~\cite{alwassel2020self,asano2020labelling} and our results, RGB alone is not sufficient to form representative clusters for videos. We tackle this problem by using \emph{OF} as complementary information to guide clustering.
\section{Method} 
\label{sec:approach}

\figref{fig:architecture} shows an overview of the proposed method. For a given video with \emph{RGB} and \emph{OF} views, we first compute two different augmentations for each view, and then pass them through the \emph{RGB} and \emph{OF} backbones, followed by a shared projection head and a shared prototype layer in order to compute cluster assignments.

\subsection{Preliminaries} \label{sec:preliminaries}
State-of-the-art clustering based SSL methods (SwAV~\cite{caron2020unsupervised}, VICC~\cite{toering2021selfsupervised},  Sela~\cite{YM.2020Self-labelling}, Selavi~\cite{asano2020self}) are built on top of the Cuturi \emph{et. al.}~\cite{cuturi2013sinkhorn} formulation. The Cuturi \emph{et. al.}~\cite{cuturi2013sinkhorn} formulation is used to find optimal assignments from samples to cluster centers (or prototypes) under constraints.

Formally, consider a given minibatch of $M$ videos $X=\{x_{1}, ..., x_{M}\}$, and $N$ prototypes $P=\{p_{1}, ..., p_{N}\}$ represented by trainable vectors. The idea of \cite{caron2020unsupervised} is to compute two random augmentations $X^{t}$ and $X^{s}$, and compute their feature vectors $F^{t}=\{f_{1}^{t}, ..., f_{i}^{t}\}$ and $F^{s}=\{f_{1}^{s}, ..., f_{i}^{s}\}$ using an encoder network $\theta$. Next, optimal assignments $D^{t}=\{d_{1}^{t}, ..., d_{M}^{t}\}$ and $D^{s}=\{d_{1}^{s}, ..., d_{M}^{s}\}$ are computed 
from features $F^{t}$ and $F^{s}$ to prototypes $P$, as described in the following paragraph. $d^{t}_{i} \in D^{t}$ and $d^{s}_{i} \in D^{s}$ vectors represent assignment values from $f^{s}_{i}$ and  $f^{t}_{i}$ to prototypes, respectively. 
Finally, $\theta$ is trained with the following loss function:
\begin{equation}
\resizebox{.5\textwidth}{!}{$
\mathcal{L}(F^{t},F^{s})=\sum_{i}^{M}\Big( l(d_{i}^{t},g_{i}^{s})+l(d_{i}^{s},g_{i}^{t})
\Big),
$}
\label{eq:l_loss_ce}
\end{equation}
where 
\begin{equation}
l(d^{t},g^{s})=-\sum_{n}^{N} d_{n}^{t}log(g^{s}_{n}),
 \hspace{2mm} 
g_{n}^{s}=\frac{exp((f^{s})^{\top}p_{n})}{\sum_{n'}^{}exp((f^{s})^{\top}p_{n'})}
\label{eq:gl}
\end{equation}
This loss function minimizes the distance of two different augmentations by comparing them according to their assignment. Ideally, the two augmentations should be assigned to the same prototype, due to identical semantic content. 

\noindent
\paragraph{Computing optimal assignment:} The optimal assignment (OA) $D$ can be found by solving the following optimization: 
\begin{equation}
d_{C}(F,P):=\min_{D \in U} \langle D,C \rangle 
\label{eq:optimal_map}
,\end{equation}
where $C \in R^{M \times N}$ is a distance matrix from a batch of feature vectors to prototypes and $\langle D,C \rangle=tr(C^{\top}D)$.
$U$ represents all possible assignments from our features $F$ to prototypes $P$. Formally: 
\begin{equation}
U:=\{D \in R_{+}^{M \times N} \hspace{1mm} | \hspace{1mm} D1_{N}=\psi, D^{\top}1_{M}=\omega \} ,\label{eq:uconstraints}
\end{equation}
where $\psi=\frac{1}{M}*\Vec{1}_{M}$ and  $\omega=\frac{1}{N}*\Vec{1}_{N}$, and $\Vec{1}_{M} \in R^{M}$ and $\Vec{1}_{N} \in R^{N}$ are ones vectors. The constraint on $U$ ensures all the prototypes are selected.  
\eqrefhc{eq:optimal_map} can be solved with linear programming. Since this is computationally expensive, it is infeasible to carry it out during training for every batch. 
Cuturi~\emph{et al}.~\cite{cuturi2013sinkhorn} address this issue by adding entropy regularization and solving the resulting problem with the Sinkhorn algorithm:
\begin{equation}
\resizebox{.5\textwidth}{!}{$
d_{C}(F,P):=\min_{D \in U} \Big( \langle D,C \rangle - \lambda_{1}h(D) \Big).
$}
\label{eq:reg_ot}
\end{equation}
where $h$ represents the entropy. The above formulation is a strictly convex and smoothed version of \eqrefhc{eq:optimal_map}. This allows us to efficiently approximate $D$.  Specifically, $D$ has a unique solution for any given $\lambda_{1}$, which is in the form of: $D^{\lambda_{1}}=\it{diag(u)}K\it{diag(v)}$
where $K=e^{\lambda_{1}C}$, and $u,v$ are non-negative unique vectors. The proof of this statement can be found in~\cite{cuturi2013sinkhorn}. According to the Sinkhorn theorem\cite{erlander1990gravity}, if $K$ has only positive elements, the unknowns $u$ and $v$ can be determined via the Sinkhorn algorithm.

There are two main problems with this conventional formulation: First, it does not allow utilizing the complementary information, which is particularly important in the video, since it has been shown that using \emph{RGB} alone encourages the model to focus on the background and ignore motion cues~\cite{asano2020labelling}. We address this problem by rewriting \eqrefhc{eq:optimal_map} such that we can fuse two information sources to obtain the cluster assignment. Second, in the conventional formulation~\cite{caron2020unsupervised,toering2021selfsupervised}, the prototypes are randomly initialized and trained with the rest of the model without any restriction, which can lead to a degenerate solution where all prototypes collapse into a single point, and consequently the feature vectors too~\cite{cai2021theory,chen2021exploring,chen2021empirical,regatti2021consensus}. To overcome this problem, we propose to use a novel prototype regularizer, which encourages prototypes to be maximally far apart. 

\subsection{Guided Online Cluster Assignment (GOCA)}
\label{sec:goca}
Our goal is to train backbone networks $\theta_{\text{RGB}}$ and $\theta_{\text{OF}}$ while sharing information between the two views. The single-view approach of~\cite{caron2018deep,caron2020unsupervised,YM.2020Self-labelling} does not allow combining the two information sources. A straightforward way is to concatenate the $F_{\text{RGB}}$ and $F_{\text{OF}}$ features, but we have observed that this approach yields a very small improvement over \emph{RGB}-only training on various downstream tasks (see Experiments \ref{prg:modal_merging}). Another way is to force the similarity of \emph{OF} and \emph{RGB} features as in~\cite{toering2021selfsupervised}. Even though this idea obtains better results on downstream tasks, it leads to a loss of information by forcing each view to only maintain mutual information. Hence, it does not reach the maximum potential performance of correctly combining the two views (\tabref{tbl:prior}). 

In contrast, we propose a principled way to combine two information sources. We use each information source as a prior to the other. We assign $D'_{\text{OF}}$ and $D'_{\text{RGB}}$ as the initial prototype assignment (computed based on \eqrefhc{eq:reg_ot}), and $D_{\text{OF}}$ and $D_{\text{RGB}}$ as the final. To this end, we use the $D'_{\text{OF}}$ as the prior to $D_{\text{RGB}}$, and $D'_{\text{RGB}}$ as the prior to $D_{\text{OF}}$

We implement this idea by modifying Eq.~\eqref{eq:reg_ot} in a way that it takes the prior into account. Our optimal assignment with prior (OAP) optimization problems for $D_{\text{RGB}}$ and $D_{\text{OF}}$ takes this form: 
\begin{equation}
\resizebox{0.9\textwidth}{!}{$
\begin{aligned}
& d_{C_{\text{RGB}}}(F_{\text{RGB}},P)
& :=\min_{D_{\text{RGB}} \in U} \Big( \langle D_{\text{RGB}},C_{\text{RGB}}\rangle  - \lambda_{1}h(D_{\text{RGB}})+ \lambda_{2}\text{KL}(D_{\text{RGB}}|D'_{\text{OF}}) \Big) \label{eq:prior_ot_rgb}
,\end{aligned} $}
\end{equation}
and
\begin{equation}
\resizebox{0.9\textwidth}{!}{$
\begin{aligned}
&d_{C_{\text{OF}}}(F_{\text{OF}},P) 
& :=\min_{D_{\text{OF}} \in U} \Big( \langle D_{\text{OF}},C_{\text{OF}}\rangle  - \lambda_{1}h(D_{\text{OF}}) + \lambda_{2}\text{KL}(D_{\text{OF}}|D'_{\text{RGB}}) \Big) \label{eq:prior_ot_flow}
.\end{aligned}
 $}
\end{equation}
% where $D'_{\text{RGB}}$ and $D'_{\text{OF}}$ are prior assignment matrices computed based on \eqrefhc{eq:reg_ot},
Where $C_{\text{RGB}}$ and $C_{\text{OF}}$ are distance matrices from \emph{RGB} and \emph{OF} features to prototypes, $\text{KL}(\cdot|\cdot)$ represents the Kullback Leibler divergence between two assignment matrices, $\lambda_{1}$ and $\lambda_{2}$ are hyper-parameters. Note that we use the same prototypes $P$ for \emph{OF} and \emph{RGB} features. These two optimization problems can be solved via the following lemma: 
\\
\\
\noindent \textbf{Lemma 1.} $D_{\text{RGB}}$ and $D_{\text{OF}}$ have a unique solution for $\lambda_{1}$ and $\lambda_{2}$ in the form of:
\begin{equation}
\begin{aligned}
D_{\text{b}}^{\lambda_{1},\lambda_{2}}=\it{diag(u_{\text{b}})}K_{\text{b}}\it{diag(v_{\text{b}})} \hspace{5mm} b \in \{RGB,OF\}
% \\
% &D_{\text{OF}}^{\lambda_{1},\lambda_{2}}=\it{diag(u_{\text{OF}})}K_{\text{OF}}\it{diag(v_{\text{OF}})}
\end{aligned}\label{eq:D_opt_problem}
\end{equation}

%Long form of wririting
\noindent \textit{Proof} We will prove only for $D_{\text{RGB}}$, but the same proof can be used for \emph{OF} as well. Let $\mathcal{L}(D_{\text{RGB}},\alpha, \beta)$ be the Lagrangian form of the Equation\eqref{eq:prior_ot_rgb} with dual variables $\alpha \in R^{M}$ and $\beta \in R^{N}$ for the two equality constraints in $U$ (see \eqrefhc{eq:uconstraints}):
\begin{equation}
\resizebox{0.9\textwidth}{!}{$
\begin{aligned}
&\mathcal{L}(D_{\text{RGB}},\alpha, \beta)= \sum_{i,j} \lambda_{1} d_{\text{RGB}_{ij}} log(d_{\text{RGB}_{ij}}) 
&+\sum_{i,j} d_{\text{RGB}_{ij}} c_{\text{RGB}_{ij}} +   \sum_{i,j} \lambda_{2} d_{\text{RGB}_{ij}} log\left(\frac{d_{\text{RGB}_{ij}}}{d'_{\text{OF}_{ij}}}\right)  \\
&+\alpha(D_{RGB}1_{M}-\psi)+\beta(D_{RGB}^{\top}1_{N}-\omega)
\end{aligned}
$}
\end{equation}
for all $(i,j)$ we set $(\partial \mathcal{L}/\partial d_{\text{RGB}_{ij}} = 0)$ and solve for $d_{\text{RGB}_{ij}}$
\begin{equation}
\resizebox{\textwidth}{!}{$
\begin{aligned}
% \begin{aligned}
% \resizebox{.35\textwidth}{!}{
% \begin{split}
d_{\text{RGB}_{ij}}=\exp\left(-0.5-\frac{\alpha_{i}}{\lambda_{1}+\lambda_{2}}\right)  \exp\left(\frac{-c_{\text{RGB}_{ij}}+\lambda_{2}log(d'_{\text{OF}_{ij}})}{\lambda_{1}+\lambda_{2}} \right)  \exp\left(-0.5-\frac{\beta_{j}}{\lambda_{1}+\lambda_{2}}\right).
% \end{split}
% }
\end{aligned}
$}
\end{equation}
When we choose $K_{\text{RGB}}=exp(\frac{-c_{\text{RGB}_{ij}}+\lambda_{2}log(d'_{\text{OF}_{ij}})}{\lambda_{1}+\lambda_{2}})$, we can see that it is strictly positive since it is the element-wise exponential, therefore according the Sinkhorn’s theorem~\cite{sinkhorn1967diagonal}, $D_{\text{RGB}}^{\lambda_1,\lambda_2}$ has unique  solution  for any given $\lambda_1$ and $\lambda_2$. Thus, $u_{\text{RGB}}$ and $v_{\text{RGB}}$ vectors in \eqrefhc{eq:D_opt_problem} can be computed with Sinkhorn algorithm. In our formulation, since the final cluster assignment relies on both views, formed clusters will be robust to noise and semantically abstract. The supplemental includes the details of the proof.

\subsection{Prototype Regularization} \label{sec:prot-reg} 
It has been shown that SSL models suffer from feature collapse, i.e. when all features are mapped to the same representation \cite{cheldualnetworks,NEURIPS2020_f3ada80d,richemond2020byol,zbontar2021barlow}. Even though this problem was partially addressed in ~\cite{caron2020unsupervised} by using equipartition constraint as in \eqrefhc{eq:optimal_map}, in practice, $\lambda$ needs to be carefully tuned~\cite{cai2021theory,chen2021exploring,chen2021empirical,regatti2021consensus}. More specifically, we observe that a higher $\lambda$ leads to numerical issues, and lower values tend to cause feature collapse. 

Alternatively, we introduce a regularization term that encourages prototypes to be maximally far apart. We achieve this by utilizing the idea of hyperspherical prototypes~\cite{mettes2019hyperspherical}. Formally, we divide the $\Phi$-dimensional hyperspherical space equally into N prototypes. For instance, for a 2-dimensional hyperspherical space (circle), this can be easily done by placing $N$ prototypes with a $\frac{2\pi}{N}$ angle difference. Even though this is easy to do for a 2-dimensional space, there is no exact solution for 3 or more dimensions~\cite{tammes1930origin}. We solve this problem by finding an approximate solution using gradient descent. Specifically, consider the $N$ prototypes that are represented with a linear layer, $W \in R^{N \times \Phi}$ in our network (\figref{fig:architecture}). Instead of training $W$ with the rest of the network end-to-end, we train it separately, once before the main training phase, by minimizing the following loss under the following constraint:
\begin{equation}
\begin{aligned}
&\mathcal{L}_{reg}=\frac{1}{N}\sum_{i}^{N}max(\Omega_{i,.}),  \hspace{2mm} \Omega =WW^{\top}-2I \hspace{2mm} s.t. \quad \forall_{i}  \lVert \mathbf{w_{i}} \rVert =1
,\end{aligned}\label{eq:prots_reg}
\end{equation}
where $I$ and $w_i$ represents the identity matrix and $i$-th row in $W$, respectively. This loss minimizes the similarity of maximally similar prototypes. To apply the constraint, we continuously re-project our prototypes to the hypersphere during training via $l2$ normalization. This can be seen as a simple and quick initialization step, which takes only 5 minutes on a Nvidia GTX 1060 GPU.
\subsection{Training procedure} \label{sec:method-concl} 
After initializing (and fixing) the prototype layer as described in \eqrefhc{eq:prots_reg}, we train the network by minimizing the following loss function: 
\begin{equation}
\mathcal{L}_{final}=\mathcal{L}_{p}(F_{\text{RGB}}^{t},F_{\text{RGB}}^{s})+\mathcal{L}_{p}(F_{\text{OF}}^{t},F_{\text{OF}}^{s})
,\end{equation}
where
\begin{equation}
\begin{aligned}
&\mathcal{L}_{p}(F_{\text{b}}^{t},F_{\text{b}}^{s})=\sum_{i}^{M}l(d_{\text{b}_i}^{t},g_{\text{b}_i}^{s}) 
+l(d_{\text{b}_i}^{s},g_{\text{b}_i}^{t}) \label{eq:curly_l_p}
\end{aligned}
\end{equation}

Here $b \in \{RGB,OF\}$ and $d_{\text{b}_i}$ $i$-th row in $D_{\text{b}}$.
$D_{\text{b}}$ represents optimal assignment matrix. We compute these matrices using \emph{Lemma 1}, while our prototypes are obtained as described in \secref{sec:prot-reg}. Note that $l$ and $g$ are defined in \eqrefhc{eq:gl}. Since \emph{RGB} and \emph{OF} features only interact when computing optimal assignment, this formulation encourages cluster similarity without enforcing \emph{RGB} and \emph{OF} features to be strictly similar, hence, not leading to information loss.
\section{Experimental Results}\label{sec:experiments}
In this section, we first describe datasets, metrics, and
training details. We present our ablation results
to show the importance of our method design choices in
\secref{sec:abl_sdy}. Then, we compare our approach with SOTA in \secref{sec:sota} for retrieval task. Then, we show our cluster analysis results in \secref{sec:cluster_analysis}. Finally, we show classification results in \secref{sec:classifcation_task}.
\\
\\
\noindent \textbf{Datasets}:
We conduct our experiments on four different datasets:
Kinetics (K400)~\cite{kay2017kinetics}, UCF~\cite{ucf101}, HMDB~\cite{hmdb}, and Diving-48~\cite{li2018resound}. We follow the same training protocols as other self-supervised learning approaches ~\cite{asano2020labelling,Han20,qian2021spatiotemporal,toering2021selfsupervised}. K400 training set contains 240k videos. UCF, HMDB, and Diving-48 contain 13k, 7k, and 17k videos respectively. 

\noindent \textbf{Evaluation Metrics}: We evaluate our model performance on the action retrieval and action classification tasks. To evaluate action retrieval, we compute \emph{Recall} at $K \in \{1,5,10,20\}$, similar to earlier works~\cite{asano2020labelling,Han20,qian2021spatiotemporal,toering2021selfsupervised}. More specifically, if the correct class is within the $K$ nearest neighbours we consider it a correct result. For action classification, we consider top-1 accuracy for \emph{Linear Classification}  and \emph{Fine Tuning} experimental setups (See \secref{prg:lin_cls}). We report all numbers in terms of percentage.

\noindent \textbf{Data Augmentations}: Following \cite{Han20,qian2021spatiotemporal,toering2021selfsupervised}, we use horizontal flipping, random cropping, Gaussian blurring, and color jittering for augmentation. We also use a \emph{multi-temporal-resolution} idea which is analogous to multi-crop in SwAV~\cite{caron2020unsupervised}. In this augmentation strategy, we sample shorter length clips alongside longer ones. We observe this makes convergence faster, but does not affect the final accuracy. 

\noindent \textbf{Backbones and Training details}: 
We conduct our experiments with two widely used backbones: S3D~\cite{xie2018rethinking} and R(2+1)D+18~\cite{tran2018closer}. We use two datasets (K400~\cite{kay2017kinetics} and UCF~\cite{ucf101}) for self-supervised training. Optical flow (\emph{OF}) is computed using the TV-L1 algorithm~\cite{zach2007duality} and pre-processed as in ~\cite{carreira2017quo,Han20,toering2021selfsupervised}. We follow earlier works~\cite{feichtenhofer2021large,qian2021spatiotemporal,toering2021selfsupervised} and use SGD+LARS~\cite{you2017large} optimizer with a learning rate of $4.8$ that is increased during the first 10 warm-up epochs and then is decreased to $0.0048$ with cosine learning rate decay. All models are pre-trained on 64 V100 GPUs (10 samples per GPU) for 500 epochs as in~\cite{asano2020labelling,feichtenhofer2021large,qian2021spatiotemporal,toering2021selfsupervised}. Each clip contains 32 consecutive frames. During the test, we turn off all augmentations and use a standard 3 (spatial) $\times$ 10 (temporal). We always denote with a $"+"$ suffix the case of \emph{RGB} and \emph{OF} being used during testing. Following the earlier works~\cite{Han20,qian2021enhancing,toering2021selfsupervised} and we use 32 and 16 frames for S3D and R(2+1)D, respectively during the
training of fine-tuning and linear-classification experiments. 

\subsection{Ablation study}\label{sec:abl_sdy}
\vspace{-3mm}
In these experiments, we demonstrate the effectiveness of each proposed component in terms of recall values. All models pre-trained on the UCF training set and evaluated on the UCF and HMDB test sets with S3D backbone. 

\noindent
\textbf{Effect of view merging strategies} \label{prg:modal_merging}
For a better understanding of the effectiveness of the proposed approach, we design 3 different baselines: \textbf{SView}, \textbf{Avg}, and \textbf{Sep}. \textbf{SView}: We train \emph{OF} and \emph{RGB} separately with identical but independent backbones, projection heads, and prototypes. This baseline can be considered a trivial extension of SwAV~\cite{caron2020unsupervised} to videos. \textbf{Avg}: We train \emph{OF} and \emph{RGB} jointly, by passing \emph{OF} and \emph{RGB} from different backbones and then feeding their average into the projection head and prototypes. \textbf{Sep}: We train the model jointly but this time \emph{OF} and \emph{RGB} information do not interact. We use separate backbones but share the projection head and prototypes. Our model differentiates from this only at the assignment computing stage. We train all our baselines by minimizing the loss described in \eqrefhc{eq:l_loss_ce}. Finally, to better compare our method (\textbf{GOCA}) with the single-view (\textbf{SView}) baseline, we also evaluate it in single-view settings by only using one of the backbones during the test. To combine \emph{RGB} and \emph{OF} representations during the test, we simply take the average of the two.

\begin{table*}
  \begin{minipage}{.49\columnwidth}
    \centering
    % \hrule
    \resizebox{\columnwidth}{!}{%
      \begin{tabular}{
			>{\raggedright\arraybackslash}p{1.4cm}
			>{\columncolor{Res2}}>{\centering\arraybackslash}p{0.7cm}
			>{\columncolor{Res2}}>{\centering\arraybackslash}p{0.7cm}
			>{\columncolor{Res1}}>{\centering\arraybackslash}p{0.8cm}
			>{\columncolor{Res1}}>{\centering\arraybackslash}p{0.8cm}
			>{\columncolor{Res3}}>{\centering\arraybackslash}p{0.8cm}
			>{\columncolor{Res3}}>{\centering\arraybackslash}p{0.8cm}} 	
	  	\multicolumn{3}{c}{} & 
		\multicolumn{2}{c}{\small {UCF}} &  
		\multicolumn{2}{c}{\small {HMDB}}\\
		\cmidrule{4-7}
		\multicolumn{1}{c}{Method} &
		\multicolumn{1}{c}{Train} &
		\multicolumn{1}{c}{Test}
		&\multicolumn{1}{c}{R@1}& \multicolumn{1}{c}{R@5}
		&\multicolumn{1}{c}{R@1}& \multicolumn{1}{c}{R@5}
		\\
	  \midrule
	 SView &   R  &  R & 41.3 & 58.9  & 17.9 & 43.2\\
	 SView &   F  &  F & 62.0 & 75.8 & 28.2 & 53.3 \\
	 VICC~\cite{toering2021selfsupervised} &  R+F  &  R & 62.1  &  77.1   & 25.5 & 49.6 \\
     VICC~\cite{toering2021selfsupervised} &  R+F  &  F & 59.7 &  77.3   & 27.7 & 53.3 \\
	 \textbf{GOCA} &  R+F  &  R & 63.4 &  76.3   & 26.3 & 50.2 \\
    \textbf{GOCA} &  R+F  &  F & 67.9 &  79.5   & 31.8 & 56.3 \\
     \midrule
     SView+ &   -  & R+F & 63.6 & 76.8  & 28.6 & 54.1\\
     Avg &   R+F  & R+F & 50.8 & 63.0  & 22.8 & 46.3\\
     Sep &   R+F  & R+F & 65.0 &  78.2  & 29.5 & 54.8 \\
      \textbf{GOCA+} &  R+F & R+F & 70.8 &  81.4   & 33.7 & 58.7\\
  
	\bottomrule
	\end{tabular}
    }
    \caption{Comparison of \emph{RGB} (R) and 
    \emph{OF} (F) merging strategies} \label{tbl:prior}
  \end{minipage}\hfill % maximize the horizontal separation
  \hspace{0.5cm}
  \begin{minipage}{.49\columnwidth}
    \centering
    % \hrule
    \vspace{2.50cm}
    \resizebox{1\columnwidth}{!}{%
     	\begin{tabular}{
         	>{\columncolor{Res7}}>{\centering\arraybackslash}p{1.2cm}
			>{\raggedright\arraybackslash}p{0.6cm}
			>{\columncolor{Res7}}>{\centering\arraybackslash}p{0.6cm}
			>{\columncolor{Res1}}>{\centering\arraybackslash}p{0.6cm}
			>{\columncolor{Res1}}>{\centering\arraybackslash}p{0.6cm}
			>{\columncolor{Res1}}>{\centering\arraybackslash}p{0.6cm}
 		    >{\columncolor{Res1}}>{\centering\arraybackslash}p{0.6cm}
     	    >{\columncolor{Res7}}>{\centering\arraybackslash}p{1.2cm}
     	    }
	  	%\toprule 	
	  	\multicolumn{3}{c}{} & 
	  	\multicolumn{3}{c}{$ \lambda_{1}$} \\  
	  	% \cmidrule{3-6}
	  	\multicolumn{3}{c}{} & 
	  	\multicolumn{1}{c}{\small {0.01}} & 
        \multicolumn{1}{c}{\small {0.02}} & 
        \multicolumn{1}{c}{\small {0.03}} & 
        \multicolumn{1}{c}{\small {0.04}} & 
        \multicolumn{1}{c}{} 
        \\
% 	  \midrule
     && 0.01 &  56.4 & 57.6 & 55.1 & 54.1 &\\
    &$ \lambda_{2}$ & 0.02 &  57.3 & 58.7 & \textbf{60.1} & 58.9 & \\
    && 0.03 &  58.1 & \textbf{60.9} & \textbf{60.5} & 57.5 & \\
     && 0.04 & 55.8 & 58.5 & \textbf{59.0} & 55.2 &\\
	\bottomrule
	\end{tabular} 
    }
    \caption{The effect of lambda parameters on recall values on the UCF RGB-only val.}\label{tbl:lambda}
  
  \end{minipage}
\end{table*}

\tabref{tbl:prior} shows our results on the UCF and HMDB test sets. In single-view settings (first 6 rows), \textbf{SView} with \emph{RGB} obtains very poor results, which can be attributed to an over-emphasis on the background scene and ignoring the motion, which is common for \emph{RGB}-only models. Our results confirm that motion-only models can perform better than \emph{RGB}, as observed in~\cite{Han20,toering2021selfsupervised}. Furthermore, GOCA outperforms both baselines even in single-view testing, which confirms that the proposed joint training approach is beneficial for each individual backbone as well.  We can also see that enforcing the similarity of the two views as in VICC does not improve \emph{OF} results compared to single-view training. In contrast, \textbf{GOCA} significantly improves the results for both views, by aligning the two representations in a smarter, more implicit manner. 

When we combine \emph{RGB} and \emph{OF} during testing (last 4 rows), we observe that naively merging features at training (\textbf{Avg}) performs better than \emph{RGB}-only by 3.5\% while significantly worse than \emph{OF}-only training (-11.2\%). This might be due to the fact that during training, \emph{RGB} information dominates the gradients, and the \emph{OF} backbone can not be fully trained. The performance of \textbf{Sep} indicates that joint training for \emph{RGB} and \emph{OF} can improve the results when we do not naively merge features.
Finally, the proposed model, \textbf{GOCA}, further improves the results, which verifies the efficacy of the proposed guided cluster assignment idea.

\noindent
\textbf{Effect of $\lambda_{1}$ and $\lambda_{2}$.}
Our next ablation study is observing the impact
of the $\lambda_{1}$ and $\lambda_{2}$ parameter values in \eqrefhc{eq:prior_ot_rgb} and \eqrefhc{eq:prior_ot_flow}. These parameters control the effect of the uniformity assumption and prior distribution. 
\tabref{tbl:lambda} shows how recall at 1 varies depending on the $\lambda_{1}$ and $\lambda_{2}$ parameters. We train GOCA for 200 epochs and evaluate it on the UCF dataset. We can observe that the proposed approach is robust to $\lambda_{1}$ and $\lambda_{2}$, especially in the range of $[0.02,0.03]$. For all other experiments, we set $\lambda_{1}=0.02$ and $\lambda_{2}=0.03$. 

  \begin{minipage}{\textwidth}
  \begin{minipage}[b]{0.45\textwidth}
  \includegraphics[scale=0.22,trim=80 140 80 0]{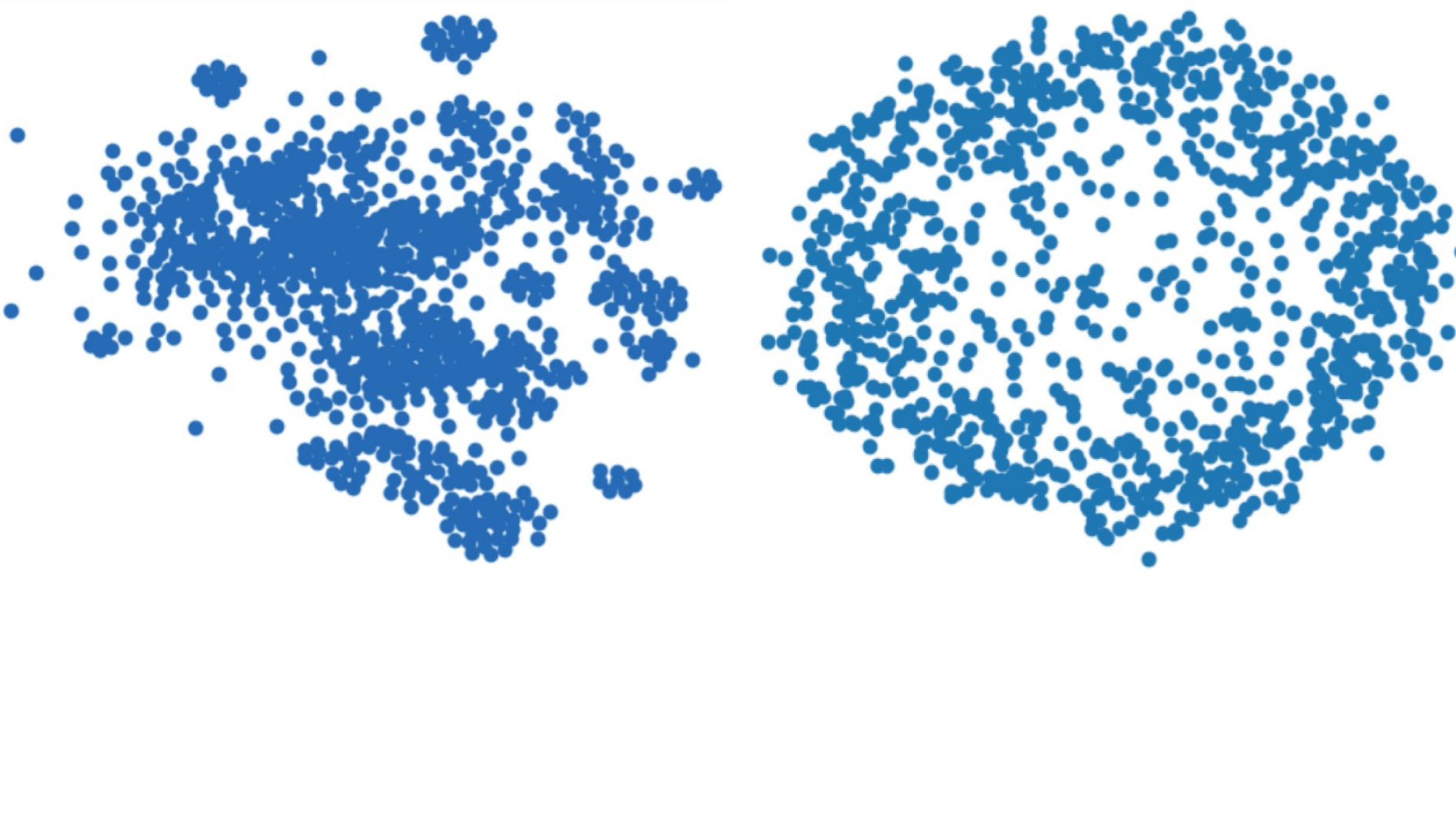}
    \captionof{figure}{The t-SNE plot of 1000 prototypes in 2D without  regularization (left) and with regularization (right)} 	\label{fig:reprot}
    \vspace{-4mm}
  \end{minipage}
  \hspace{0.5cm}
%   \hfill
  \begin{minipage}[b]{0.45\textwidth}
    % \centering
\resizebox{.90\textwidth}{!}{
\begin{tabular}{
			>{\raggedright\arraybackslash}p{1.2cm}
			>{\raggedright\arraybackslash}p{1cm}
			>{\columncolor{Res2}}>{\centering\arraybackslash}p{0.75cm}
			>{\columncolor{Res2}}>{\centering\arraybackslash}p{0.6cm}
			>{\columncolor{Res1}}>{\centering\arraybackslash}p{0.8cm}
			>{\columncolor{Res1}}>{\centering\arraybackslash}p{0.8cm}
			>{\columncolor{Res3}}>{\centering\arraybackslash}p{0.8cm}
			>{\columncolor{Res3}}>{\centering\arraybackslash}p{0.8cm}}
	  	%\toprule 	
	  	\multicolumn{1}{c}{} & 
	  	\multicolumn{1}{c}{Use} & 
	  	\multicolumn{2}{c}{} & 
		\multicolumn{2}{c}{\small {UCF}} &  
		\multicolumn{2}{c}{\small {HMDB}}\\
		\cmidrule{5-8}
		\multicolumn{1}{c}{Method} &
		\multicolumn{1}{c}{ProtReg} &
		\multicolumn{1}{c}{Train}&
		\multicolumn{1}{c}{Test}
		&\multicolumn{1}{c}{R@1}& \multicolumn{1}{c}{R@5}
		&\multicolumn{1}{c}{R@1}& \multicolumn{1}{c}{R@5}\\
	  \midrule
       SView&\textbf{No} &     R & R &  40.9 & 57.4 & 17.1 & 42.7\\
       SView&\textbf{Yes} &     R & R & 42.3 & 58.9  & 17.9 & 43.2 \\
      \midrule
    %  \textbf{No} &   F &  59.0 & 74.4  & 27.4 & 52.5 \\
    %  \textbf{Yes} &F & 62.0 & 75.8  & 28.2 & 53.3 \\
    %  \midrule
	\textbf{GOCA+}& \textbf{No} &     R+F & R+F &  69.1 & 79.4 & 32.9 & 58.6 \\
       \textbf{GOCA+}&\textbf{Yes} &    R+F & R+F &  70.8 &  81.4 & 33.6 & 58.7  \\
	
	\bottomrule
	\end{tabular}
		}
	  \captionof{table}{Effect of prototype  regularization on recall values} \label{tbl:protreg}
    \end{minipage}
\vspace{10mm}
  \end{minipage}
\noindent
\textbf{Effect of prototype regularizer.} 
 Our prototype regularizer idea guarantees that prototypes are maximally far apart from each other. \figref{fig:reprot} shows t-SNE plot of 1000 prototypes with and without regularization. As we can see, the proposed method locates the prototypes maximally far apart from each other, on the other hand, in the unregularized case, prototypes are quite closely grouped. As we discuss in \secref{sec:prot-reg}, maximally far apart prototypes prevent the feature collapse and allow us stable training. Another benefit is that clusters formed around prototypes are also will be far apart from each other which is particularly important for retrieval tasks. We verify this benefit by comparing the recall accuracy with and without using prototype regularization. Tab. \tabref{tbl:protreg} shows the consistent effectiveness of the proposed method for both models on both datasets. 
% \FloatBarrier
\begin{table*}[!ht]
    \vspace{-7mm}
	 \addtolength{\tabcolsep}{-2.0pt}    

	\centering{
		\resizebox{.80\textwidth}{!}{
	\begin{tabular}{
			>{\raggedright\arraybackslash}p{2.2cm}
			>{\columncolor{Res2}}>{\centering\arraybackslash}p{1.2cm}
			>{\columncolor{Res2}}>{\centering\arraybackslash}p{2.3cm}
			>{\columncolor{Res2}}>{\centering\arraybackslash}p{0.9cm}
			>{\columncolor{Res2}}>{\centering\arraybackslash}p{1.1cm}
			>{\columncolor{Res1}}>{\centering\arraybackslash}p{0.9cm}
			>{\columncolor{Res1}}>{\centering\arraybackslash}p{0.9cm}
			>{\columncolor{Res1}}>{\centering\arraybackslash}p{0.9cm}
			>{\columncolor{Res1}}>{\centering\arraybackslash}p{0.9cm}
			>{\columncolor{Res3}}>{\centering\arraybackslash}p{0.9cm}
			>{\columncolor{Res3}}>{\centering\arraybackslash}p{0.9cm}
			>{\columncolor{Res3}}>{\centering\arraybackslash}p{0.9cm}
			>{\columncolor{Res3}}>{\centering\arraybackslash}p{0.9cm}}
	  	%\toprule 	
	  	\multicolumn{5}{c}{} & 
		\multicolumn{4}{c}{\small {UCF}} &  
		\multicolumn{4}{c}{\small {HMDB}}\\
		\cmidrule{6-13}
		\multicolumn{1}{c}{Method} &
		\multicolumn{1}{c}{DS} &  
		\multicolumn{1}{c}{Backbone}  &  
		\multicolumn{1}{c}{Res}  &  
		\multicolumn{1}{c}{Modl}
		&\multicolumn{1}{c}{R@1}& \multicolumn{1}{c}{R@5}& \multicolumn{1}{c}{R@10}&  \multicolumn{1}{c}{R@20}
		&\multicolumn{1}{c}{R@1}& \multicolumn{1}{c}{R@5}& \multicolumn{1}{c}{R@10}& \multicolumn{1}{c}{R@20}\\
	  \midrule
	  Selavi~\cite{asano2020labelling} & K400 & R(2+1)D+18 & $112$   & V+A &   52.0 &  68.6 & -  &  84.5 &  24.8 &   47.6 &   -  &    75.5  \\

	  Rbst-xID~\cite{morgado2021robust} &  K400 &R(2+1)D+18& $112$   & V+A &  60.9 &   79.4 &  - &  \textbf{90.8} &  30.8 &   55.8&  - &   79.7  \\
	  TCGL~\cite{Liu9713748} &  K400 &R(2+1)D+18& $112$   & V &  21.5 &   39.3 &  49.3 &  59.5 &  10.5 &  27.6&  39.7 &   55.63  \\
	  MotionFit~\cite{gavrilyuk2021motion} &  K400 &R(2+1)D+18& $112$   & V &  61.6 &   75.6 &  - &85.5 &  29.4  & 46.5&   - & 66.7  \\
	  ASCNet~\cite{Huang_2021_ICCV} &  K400 &R3D-18& $112$   & V &  58.9 &    76.3 &  82.2 &   87.5 & - &     -&   - & -  \\
	  
	  Enhenced~\cite{qian2021enhancing} &  K400 &R3D-18& $112$   & V &  41.5 &   60.6 &  71.2 &   80.1 &  20.7 &     40.8&   55.2 & 68.3  \\

	   MCN~\cite{lin2021self} &  K400 &R(2+1)D-18& $128$   & V &  52.5 &    69.5 & 77.9 &   83.1 & 23.7 &     46.5&   58.9& 72.4 \\
	  
	   Zhang~\cite{lin2021dualvar} &  K400 &R3D-18& $112$   & V &  46.7 &   63.1 &  69.7 &78.0 &  - &     -&   - & -  \\
	  CoCLR~\cite{Han20} &  K400 &S3D & 128   & V &  45.6 &  63.9 & 75.4 &  81.7 &  - &  - &   - &    -  \\
	
      \textbf{GOCA} &  K400 &S3D & 128   & V &  67.3 & 79.1 & 84.9 & 89.9 & 32.7 & 55.1 &  68.5 &  79.5\\
      \textbf{GOCA+} &  K400 &S3D & 128   & V &  \textbf{68.6} &  \textbf{80.7}  &  \textbf{86.6} & \textbf{90.1} & \textbf{33.2} & \textbf{56.3} &  \textbf{68.5} &  \textbf{80.2} \\
 	  \midrule
	  \midrule
	  VCOP~\cite{xu2019self} & UCF & R(2+1)D+18 & $112$   & V &   14.1 &  30.3 & 40.4 &  51.1 &  7.6 &  22.9 & 34.4  & 48.8  \\
	  
 	  Var. PSP~\cite{cho2021self} & UCF & R(2+1)D+18 & $112$   & V &  24.6 &  41.9 & 51.3 &  62.7 &  - & - & -  & -  \\
	  RTT~\cite{jenni2020video} & UCF & R(2+1)D+18 & $112$   & V &  26.1 &  48.5 & 59.1 & 69.6  &  - & - & -  & -  \\
	  
	  MemDPC~\cite{han2020memory} & UCF & R2D3D-18 & $112$   & V &  20.2 &  40.4 &  52.4 &  64.7  & 7.7 & 25.7& 40.6  & 57.7  \\
	  
	  VCP~\cite{luo2020video} & UCF & R(2+1)D+18 & 112   & V & 19.9 &  33.7 &  42.0 & 50.5  & 6.7 & 21.3& 32.7  & 49.2 \\

       TCLR~\cite{dave2021tclr} & UCF & R(2+1)D+18 & $112$   & V &   56.9 &  72.2 & 79.0 &  84.6 &  24.1 &  45.8 & 58.3  & 75.3  \\
	  Enhenced~\cite{qian2021enhancing} &  UCF &R3D-18& $112$   & V &  39.6 &   57.6 &  69.2 &  78.0 &  18.8 &    39.2&  51.0 &  63.7  \\
	   
	   \textbf{GOCA} &  UCF &R(2+1)D+18  & 112   & V &  62.8 & 77.7 & 82.0 & \textbf{87.0} & 22.3 & 47.2 &  60.1 &  73.3\\
      \textbf{GOCA+} &  UCF &R(2+1)D+18  & 112   & V &  \textbf{63.4} &  \textbf{78.6}  &  \textbf{82.5} & 86.2 & \textbf{28.5} & \textbf{54.4} &  \textbf{66.2} &  \textbf{76.6} \\
	   \midrule
	  \textcolor{lightgray}{MCL+~\cite{li2021motion}} &  \textcolor{lightgray}{UCF} &\textcolor{lightgray}{S3D} & \textcolor{lightgray}{224}   & \textcolor{lightgray}{V} &  \textcolor{lightgray}{67.0} &   \textcolor{lightgray}{80.8} & \textcolor{lightgray}{86.3} &  \textcolor{lightgray}{90.8} &  \textcolor{lightgray}{26.7} &   \textcolor{lightgray}{52.5}&   \textcolor{lightgray}{67.0} &   \textcolor{lightgray}{79.3}  \\
	   Time-Equ*~\cite{jenni2021time} &  UCF &R3D-18 & 128   & V &  62.1 &  - &  - &  - &  31.5 &    -&  - &     -  \\
	  CoCLR~\cite{Han20} &  UCF &S3D & 128   & V &  53.3 &  69.4 & 76.6 &  82.0 &  23.2 &   43.2&   53.5 &    65.5  \\
	  CoCLR+~\cite{Han20} &  UCF &S3D & 128    & V &  55.9 &   70.8 &  76.9 &  82.5 &  26.1 &   45.8&   57.9 &   69.7  \\

	   ViCC~\cite{toering2021selfsupervised} &  UCF &S3D & 128    & V &  62.1 & 77.1 & 83.7 & 87.9 & 25.5 & 49.6 &  61.9 &  72.5 \\
      ViCC+~\cite{toering2021selfsupervised} &  UCF &S3D & 128    & V &  65.1 &  80.2  &  \textbf{85.4} & \textbf{89.8} & 29.7& 54.6 &  66.0 &  76.2 \\

      \textbf{GOCA} &  UCF &S3D & 128    & V &  63.4 &  76.3  &  81.3 & 86.5 & 26.3 & 50.2 & 62.6 &  77.0\\
      \textbf{GOCA+} &  UCF &S3D & 128    & V &  \textbf{70.8} &  \textbf{81.4}  &  85.3 & 89.5 & \textbf{33.7} & \textbf{58.5} &  \textbf{70.0} &  \textbf{80.6}  \\
	\bottomrule
	\end{tabular}}}
	\caption{
	Video retrieval results on UCF~\cite{ucf101} and HMDB51~\cite{hmdb} datasets. \emph{DS}, \emph{Res}, and \emph{Modl} represent training dataset, input resolution, and input modality respectively. Rows above the double line are trained on Kinetics and the rest use UCF.  Light grey colored methods (MCL~\cite{li2021motion}) use 2 times more input resolution (224). Time-Equ*~\cite{jenni2021time} uses various additional loss functions that used in earlier works (speed~\cite{benaim2020speednet,Yao_2020_CVPR}, direction~\cite{pickup2014seeing}, order~\cite{xu2019self}). These loss functions can be combined with our loss functions as well.
}\label{tbl:knn}
\end{table*}

\noindent
\textbf{Comparison with the state-of-the-art} %\label{sec:sota}
Given the large body of self-supervised video understanding 
works published thus far, we only selected recent (from 2019) publications. To the best of our ability, we conducted a fair comparison. However, we still observe small variations in terms of input resolution and fine-tuning details in the literature, which makes it extremely hard to perform perfectly fair comparisons. Furthermore, we encourage the readers to study the supplementary material for more in-depth experimental results. As noted in \cite{feichtenhofer2021large}, we observe that during the \emph{Fine-Tuning} experiment, backbone networks tend to overfit validation datasets (UCF and HMDB), therefore we perform extensive retrieval and cluster analysis experiments to show our contributions' influence (On these experiments there is no supervised training on validation datasets). 

\begin{figure}[!ht]
\centering
 \includegraphics[clip,trim=0.1cm 2.0cm 0cm 2.3cm,width=0.65\linewidth]{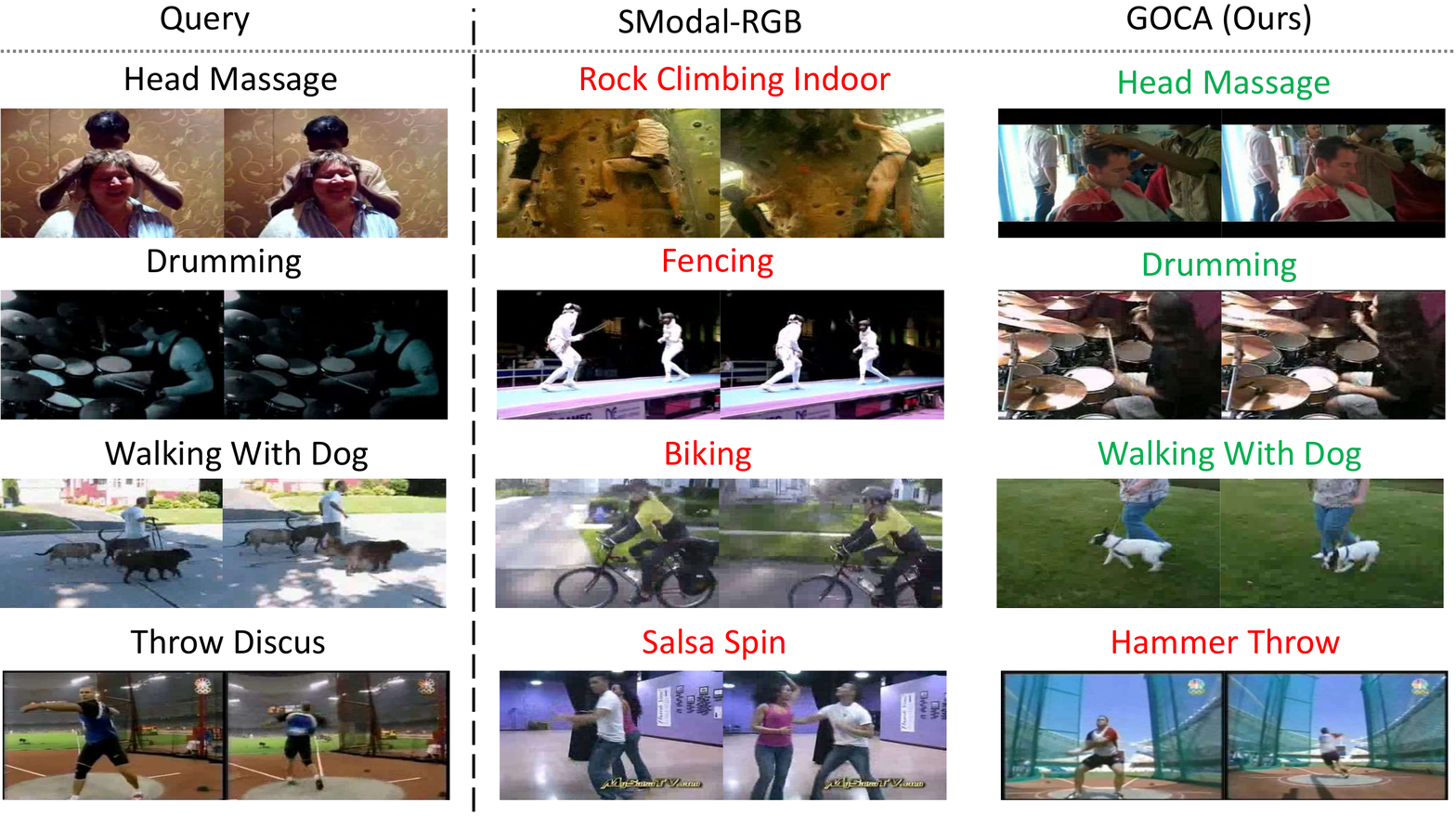}
 \vspace{-4mm}
  \caption{Retrieval comparison between SView-RGB and GOCA representations on UCF dataset. Query videos (left) are selected from test set and top-1 nearest neighbour videos are retrieved from the training set. Ground truth labels are reported on top of each query.}
  \label{fig:retrievals_ucf_goca}
\end{figure}

\begin{table}[!ht]
	\centering{
		\resizebox{.60\textwidth}{!}{
	\begin{tabular}{
			>{\raggedright\arraybackslash}p{4.0cm}
			>{\columncolor{Res1}}>{\centering\arraybackslash}p{1.5cm}
			>{\columncolor{Res1}}>{\centering\arraybackslash}p{1.5cm}
		    >{\columncolor{Res1}}>{\centering\arraybackslash}p{1.5cm}
		    >{\columncolor{Res1}}>{\centering\arraybackslash}p{1.5cm}
		    }
	  	%\toprule 	
	  	\multicolumn{1}{c}{Method} & 
  		\multicolumn{1}{c}{R@1}& 
  		\multicolumn{1}{c}{R@5}& 
  		\multicolumn{1}{c}{R@10}&  
  		\multicolumn{1}{c}{R@20}\\
	  \midrule
      CoCLR~\cite{Han20} &  6.7 &23.6 & 37.4&55.2  \\
      VICC~\cite{toering2021selfsupervised} & 7.2&24.4&38.1&54.5   \\
     \textbf{GOCA} &   \textbf{8.1} & \textbf{24.8} & \textbf{38.3}&55.1   \\
	
	\bottomrule
	\end{tabular}}}
	\caption{Retrieval results on Diving-48~\cite{li2018resound} with S3D backbone(pre-trained on UCF) }
		\label{tbl:dive_48}
\end{table}

\begin{table}[!ht]
	\centering{
		\resizebox{.60\textwidth}{!}{
	\begin{tabular}{
			>{\raggedright\arraybackslash}p{4.2cm}
			>{\columncolor{Res1}}>{\centering\arraybackslash}p{2.5cm}
			>{\columncolor{Res1}}>{\centering\arraybackslash}p{2.5cm}
			>{\columncolor{Res1}}>{\centering\arraybackslash}p{2.5cm}}
	  	%\toprule 	
	  	\multicolumn{1}{c}{Method} & 
	  	\multicolumn{1}{c}{\small {Acc}} &  
		\multicolumn{1}{c}{\small {NMI}} &  
		\multicolumn{1}{c}{\small {F1}}\\
	  \midrule
	  SView-RGB &    42.1 & 65.1 &40.3  \\
      CoCLR~\cite{Han20} &    44.3 &65.4 & 41.9 \\
      CoCLR+~\cite{Han20} &  46.1 &68.6 & 43.1  \\
      VICC~\cite{toering2021selfsupervised} &    51.1 &70.4 & 50.0  \\
      VICC+~\cite{toering2021selfsupervised} & 51.9&72.9&53.7   \\
	 \textbf{GOCA} &   57.3 & 75.8 & 57.4  \\
     \textbf{GOCA+} &   \textbf{61.2} & \textbf{78.7} & \textbf{61.3}   \\
	
	\bottomrule
	\end{tabular}}}
	\caption{Cluster quality results on UCF test set with S3D backbone. All models are trained on UCF training set.}
		\label{tbl:cluster_quality}
		\vspace{-5mm}
\end{table}

\begin{table*}
  \begin{minipage}{.40\columnwidth}
    \centering
    % \hrule
    \resizebox{\columnwidth}{!}{%
     	\begin{tabular}{
			>{\raggedright\arraybackslash}p{2cm}
			>{\columncolor{Res2}}>{\centering\arraybackslash}p{0.7cm}
			>{\columncolor{Res2}}>{\centering\arraybackslash}p{1.7cm}
			>{\columncolor{Res2}}>{\centering\arraybackslash}p{0.5cm}
			>{\columncolor{Res1}}>{\centering\arraybackslash}p{0.7cm}
			>{\columncolor{Res3}}>{\centering\arraybackslash}p{1.1cm}}
	    
	   % \multicolumn{6}{c}{\textbf{Linear Classification}} \\
	    
	    \midrule
		\multicolumn{1}{c}{Method}
		&\multicolumn{1}{c}{DS} 
		&\multicolumn{1}{c}{Backbone} 
		&\multicolumn{1}{c}{Res} 
		&\multicolumn{1}{c}{UCF}
		&\multicolumn{1}{c}{HMDB}\\
	   \textcolor{lightgray}{CVRL~\cite{qian2021spatiotemporal}} & \textcolor{lightgray}{K400} & \textcolor{lightgray}{R3D-50} & \textcolor{lightgray}{224}   & \textcolor{lightgray}{92.9} & \textcolor{lightgray}{67.9}\\
	   %MoCo~\cite{feichtenhofer2021large} & K400 &R-50 & 224    & 93.3 & - \\
	   \textcolor{lightgray}{MoCo~\cite{feichtenhofer2021large}} & \textcolor{lightgray}{K400} &\textcolor{lightgray}{R-50} & \textcolor{lightgray}{224}     & \textcolor{lightgray}{93.2} & \textcolor{lightgray}{70.6} \\
	   %TaCo~\cite{bai2020can} & K400 & R50 & 224    & 85.1 & 51.6\\
	  \hdashline[5pt/5pt]
	   ASCNet\cite{qian2021enhancing} & K400 & R3D-18 & 112    & 80.5 &  52.3\\
	   Enhanced\cite{qian2021enhancing} & K400 & R3D-18 & 112    & 79.1 &  47.6\\
	   CoCLR~\cite{Han20} & K400 & S3D & 128    & 87.9 & 54.6\\
	   CoCLR+~\cite{Han20} & K400& S3D & 128  & 90.6 &
	   62.9\\

	   \textbf{GOCA} & K400 & S3D & 128    & 89.3 & 63.2\\
	   \textbf{GOCA+} & K400 & S3D & 128    & \textbf{91.1} &
	   \textbf{65.8}\\
	   \midrule
	   \midrule
	   MoCo*~\cite{feichtenhofer2021large} & UCF &R(2+1)D+18 & 112& 77.6 & 45.7 \\
	   %VCOP~\cite{xu2019self} & UCF & R(2+1)D+18 & 112    &  72.4 &  30.9\\
	   CVRL*~\cite{qian2021spatiotemporal} & UCF & R3D+18 & 112   & 75.8 &44.6\\
	   TCLR~\cite{dave2021tclr} & UCF & R(2+1)D+18 & 112   & 82.4 &52.9\\
	   TCLR~\cite{dave2021tclr} & UCF & R(2+1)D+18 & 112   & 82.8 &55.6\\
	   
	   Var. PSP~\cite{cho2021self} & UCF & R3D+18 & 112    &  74.8 &  36.8\\
	      
	   PacePred~\cite{wang2020self} & UCF & R(2+1)D+18 & 112     &  75.9 & 35.9\\
	   VCP~\cite{luo2020video} & UCF & R(2+1)D+18 & 112    &  66.3 & 32.2\\
	   PRP~\cite{Yao_2020_CVPR} & UCF1 & R(2+1)D+18 & 112    & 72.1 & 35.0\\
	   RTT~\cite{jenni2020video} & UCF & R(2+1)D+18 & 112    & 81.6 & 46.4\\
	   Enhanced~\cite{qian2021enhancing} & UCF & R3D+18 & 112    & 76.2 &  41.1\\
	   Zhang~\cite{lin2021dualvar} & UCF & R(2+1)D+18 & 112    & 79.0 &  45.4\\
	   
	   \textbf{GOCA} & UCF & R(2+1)D+18 & 112     &  82.1 &   54.7\\
	   \textbf{GOCA+} & UCF & R(2+1)D+18 & 112    & \textbf{88.7} &  \textbf{60.1} \\
	 
	  \midrule

	   CoCLR~\cite{Han20} & UCF & S3D & 128   & 81.4 & 52.1\\
	   CoCLR+~\cite{Han20} & UCF & S3D & 128    & 87.3 &  58.7\\
	   ViCC~\cite{toering2021selfsupervised} & UCF & S3D & 128    &  84.3 &   47.9\\
	   ViCC+~\cite{toering2021selfsupervised} & UCF & S3D & 128    & \textbf{90.5} &   62.2\\
	   
	   \textbf{GOCA} & UCF & S3D & 128     &  83.4 &   53.5\\
	   \textbf{GOCA+} & UCF & S3D & 128  & 90.2 &  \textbf{64.8}\\
	\bottomrule
	\end{tabular}
    }
    \caption{
% 	\cHu{
	Fine Tuning. Light grey colored methods use much higher input resolution. MoCo* and CVRL* results obtained from \cite{lin2021dualvar} and \cite{dave2021tclr}, respectivelly.
% 	}
	}\label{tbl:lc_merge}
  \end{minipage}
%   \hfill % maximize the horizontal separation
  \hspace{0.5cm}
  \begin{minipage}{.40\columnwidth}
    \centering
    % \hrule
    \vspace{1.2cm}
    \resizebox{1\columnwidth}{!}{%
     	\begin{tabular}{
			>{\raggedright\arraybackslash}p{2.1cm}
			>{\columncolor{Res2}}>{\centering\arraybackslash}p{0.7cm}
			>{\columncolor{Res2}}>{\centering\arraybackslash}p{1.7cm}
			>{\columncolor{Res2}}>{\centering\arraybackslash}p{0.5cm}
			>{\columncolor{Res1}}>{\centering\arraybackslash}p{0.7cm}
			>{\columncolor{Res3}}>{\centering\arraybackslash}p{1.1cm}}
	    
	   % \multicolumn{6}{c}{\textbf{Linear Classification}} \\
	    
	    \midrule
		\multicolumn{1}{c}{Method}
		&\multicolumn{1}{c}{DS} 
		&\multicolumn{1}{c}{Backbone} 
		&\multicolumn{1}{c}{Res} 
		&\multicolumn{1}{c}{UCF}
		&\multicolumn{1}{c}{HMDB}\\
       
	  \midrule
	   \textcolor{lightgray}{MemDPC~\cite{han2020memory}} & \textcolor{lightgray}{K400} & \textcolor{lightgray}{R-2D3D} & \textcolor{lightgray}{224}    & \textcolor{lightgray}{54.1} &\textcolor{lightgray}{30.5}\\
	    \textcolor{lightgray}{CVRL~\cite{qian2021spatiotemporal}} & \textcolor{lightgray}{K400} & \textcolor{lightgray}{R3D-50} & \textcolor{lightgray}{224} & \textcolor{lightgray}{89.8} &\textcolor{lightgray}{58.3}\\
	   \hdashline[5pt/5pt]
	   TCLR~\cite{dave2021tclr} & UCF & R3D+18 & 112   & 69.9 &52.8\\
	   Enhanced\cite{qian2021enhancing} & K400 & R3D+18&112   & 63.2 &33.4\\
	   CoCLR~\cite{Han20} & K400 & S3D& 128    & 74.5 &46.1\\
	   CoCLR+~\cite{Han20} & K400 & S3D& 128    & 77.8  &52.4\\
	   \textbf{GOCA} & K400 & S3D& 128 & 78.9  &50.5\\
	   \textbf{GOCA+} & K400 & S3D& 128   &\textbf{82.8}  &\textbf{58.7}\\
	    \midrule
        \midrule
     
	   \textcolor{lightgray}{MCL+~\cite{li2021motion}} & \textcolor{lightgray}{UCF} & \textcolor{lightgray}{S3D}& \textcolor{lightgray}{224}    & \textcolor{lightgray}{79.8}  &\textcolor{lightgray}{-}\\
	    Time-Equ~\cite{jenni2021time} & UCF & R3D-18& 128    & 74.1  & 47.5\\
	   CoCLR~\cite{Han20} & UCF & S3D& 128    & 70.2  &39.1\\
	   
	   CoCLR+~\cite{Han20} & UCF & S3D& 128    & 72.1  &40.2\\
	   ViCC~\cite{toering2021selfsupervised} & UCF & S3D& 128     & 72.2  &38.5\\
	   ViCC+~\cite{toering2021selfsupervised} & UCF & S3D& 128    & 78.0  &47.9\\
	   
	   \textbf{GOCA} & UCF & S3D& 128    & 69.2  &38.6\\
	   \textbf{GOCA+} & UCF & S3D& 128   & \textbf{81.1}  &\textbf{50.0}\\
	\bottomrule
	\end{tabular}
    }
	 \caption{
% 	 \cHu{
	 Linear Classification.
% 	 }
	 }\label{tbl:ft_merge}
    
  \end{minipage}
\vspace{-5mm}
\end{table*}

\noindent
\subsection{Retrieval Results}\label{sec:sota}
% \paragraph{Retrieval Results}
\tabref{tbl:knn} and \tabref{tbl:dive_48} show the retrieval results for $Recall@K$ ($R@K$). In this experiment, we follow the standard protocol defined in~\cite{dave2021tclr,han2020memory,Han20,jenni2020video,luo2020video,qian2021enhancing,toering2021selfsupervised} and perform the evaluation on the frozen features that were computed from a pre-trained model. At Tab. \tabref{tbl:knn}, the first part of the table (above the double line) includes models that are all pre-trained on K400~\cite{kay2017kinetics}, while evaluated on the UCF and HMDB datasets. We can see a significant improvement (5\% and 8\% on UCF and HMDB respectively, averaged over all $K$ values). Surprisingly, the proposed approach even outperforms models~\cite{asano2020labelling,morgado2021robust} that use an additional view (Audio) by a large margin. Below the double line, we show models that are pre-trained on the UCF training set, using two different backbone architectures, S3D and $R(2+1)D$, separated by a horizontal line. On the S3D, we achieve an improvement of 5.7\% on UCF and 4.0\%
on HMDB at $K=1$. On the $R(2+1)D$, we can see a nearly 6\% increase across all recall values for UCF and 5\% for HMDB. Our \emph{RGB}-only model also achieves state-of-the-art results and outperforms baselines by a large margin on both datasets for both backbones. These findings verify our intuition that effectively utilizing OF can result in strong spatio-temporal representations.  We observe that 
Time-Equ*~\cite{jenni2021time} method also performs well however, this work uses speed~\cite{benaim2020speednet,Yao_2020_CVPR}, direction~\cite{pickup2014seeing}, and order~\cite{xu2019self} as auxiliary loss as well. These loss functions can be combined with our method also. Time-Equ*~\cite{jenni2021time} obtains 52.1\% and 21.4\% at at $K=1$ on  UCF and HMDB respectively, without these auxiliary loss function. We also evaluate our pre-trained model on  motion centric dataset~\cite{li2018resound}.  Tab. \ref{tbl:dive_48} show our result for Diving-48~\cite{li2018resound}. We can see that proposed model improves the baselines for the  motion centric dataset as well and obtain SOTA results. 
\figref{fig:retrievals_ucf_goca} shows our retrievals results. GOCA retrieves videos from the same semantic categories and it fails only for one case where it confuses \emph{Disk Throwing} with \emph{Hammer Throwing}. This is a quite hard example because these two activities have very similar motions and the only difference is the quite small object in the person's hand. Higher resolution input images would help to solve this problem. We can see that SModel-RGB fails to retrieve relevant videos and consistently retrieves based on the background.

\subsection{Cluster Analysis}\label{sec:cluster_analysis}

Ideally, video representations should form semantic clusters in order to facilitate activity recognition. To evaluate the clustering quality of our learned representations, we use two metrics that measure the correlation between clusters and semantic labels (ground truth). To this end, we perform K-means clustering on the our representations that are extracted from the UCF101 test set. Then we use the given labels of the test set to determine a label for each cluster via majority voting. We assign each cluster's label to all its members and compare those assigned labels to given labels. Specifically, we compute the accuracy of the assigned labels (Acc), as well as the $F_{1}$ score, which is the harmonic mean of recall and precision. We also compute the Normalized Mutual Information (NMI), which measures the mutual information between clusters and ground truth labels, divided by the sum of their entropy. Due to the randomness of K-means, we repeat experiments 50 times and take the average for each metric. As shown in \tabref{tbl:cluster_quality}, GOCA significantly improves the cluster quality in terms of all metrics, both with and without \emph{OF}. This verifies the high quality semantic clustering ability of our method.

\subsection{Classification Tasks}\label{sec:classifcation_task}
\noindent
\paragraph{Linear Classification} \label{prg:lin_cls} 
% We follow the earlier %
We follow the earlier works of \cite{Han20,qian2021enhancing,toering2021selfsupervised} for the linear classification experiments. 
After the self-supervised training on the K400 dataset, we discard the projection head and prototypes and replace them with a linear layer. Then we train the linear layer on the training set of each downstream dataset (UCF and HMDB) with frozen backbone. 
The results are shown in the first section of \tabref{tbl:lc_merge}, and demonstrate that the proposed model significantly outperforms CoCLR~\cite{Han20} on both datasets by 5\% and 6\%, respectively. 
Notably, when we combine \emph{RGB} and \emph{OF}, we achieve state-of-the-art results on the HMDB dataset, even though the other methods~\cite{han2020memory,qian2021spatiotemporal} benefit from a higher input resolution. For the case of UCF dataset, our model with the S3D marginally outperforms the other methods, achieving 80.1\% on UCF and 50.0\% on HMDB. 

\paragraph{Fine-Tuning}\label{prg:ft_cls} 
We follow the standard protocol from \cite{Han20,qian2021enhancing,toering2021selfsupervised}, where we train the full backbone on the downstream tasks. We summarize the results in the second section of \tabref{tbl:ft_merge}. On the S3D backbone, our method improves CoCLR+~\cite{Han20} by 0.5\% and 2.9\%, respectively. In addition, when using only \emph{RGB} as input, we achieve improvements of 1.4\% and 8.8\% on both datasets. 
Moving to the UCF dataset, our approach obtains state-of-art results with R(2+1)D; while, with S3D, we are slightly worse than ViCC+~\cite{toering2021selfsupervised}. 
However, on the HMDB dataset, our results outperform ViCC+~\cite{toering2021selfsupervised} by a large margin of 6.6\% and 2.6\%  when using \emph{RGB} and \emph{RGB}+\emph{OF}. Note that CVRL~\cite{qian2021spatiotemporal} and $\rho$ BYOL~\cite{feichtenhofer2021large} use higher input resolution (224) than ours (112), which leads to better performance, but needs significantly more computation. In fact, when we compare to CVRL at a resolution of 112, we significantly outperform CVRL~\cite{qian2021spatiotemporal}.

\section{Conclusion}    \label{sec:conclusion}
In this paper, we presented a novel self-supervised learning approach for videos. We showed that the proposed guided online clustering idea and the prototype regularization approach both substantially improve the performance of our learned representations on both activity retrieval and action classification tasks. We believe that our work establishes a new direction for SSL research on multi-view and multi-modal data. Although we conduct our experiments using \emph{RGB} and optical flow, the proposed idea can be applied to fuse other modalities such as \emph{RGB}+\emph{Audio} or \emph{RGB}+\emph{Text}, which we will explore in our future work. Furthermore, our method simplifies the training procedure for multi-modal SSL on videos (CoCLR and VICC require multi-stage training).

% \input{acknowledgement}
% \clearpage
% ---- Bibliography ----
%
% BibTeX users should specify bibliography style 'splncs04'.
% References will then be sorted and formatted in the correct style.
%
\bibliographystyle{splncs04}
\bibliography{egbib}
\end{document}